\documentclass{article}

\PassOptionsToPackage{numbers, compress}{natbib}

\usepackage[final]{neurips_2018}

\usepackage[utf8]{inputenc} % allow utf-8 input
\usepackage[T1]{fontenc}    % use 8-bit T1 fonts
\usepackage{hyperref}       % hyperlinks
\usepackage{url}            % simple URL typesetting
\usepackage{booktabs}       % professional-quality tables
\usepackage{amsfonts}       % blackboard math symbols
\usepackage{nicefrac}       % compact symbols for 1/2, etc.
\usepackage{microtype}      % microtypography
\usepackage{graphicx}
\usepackage{color}
\usepackage{subcaption}
\usepackage{multirow}
\usepackage{tikz}
\usepackage{amsmath}        % use \text in equation mode 

\newcommand{\tikzcircle}[2][red,fill=red]{\tikz[baseline=-0.5ex]\draw[#1,radius=#2] (0,0) circle ;}

\newcommand{\bigO}[1]{\ensuremath{\mathop{}\mathopen{}\mathcal{O}\mathopen{}\left(#1\right)}}

\title{Neural Code Comprehension: A Learnable Representation of Code Semantics}

\author{
  Tal Ben-Nun\\
  ETH Zurich\\
  Zurich 8092, Switzerland\\
  \texttt{talbn@inf.ethz.ch}
  \And Alice Shoshana Jakobovits\\
  ETH Zurich\\
  Zurich 8092, Switzerland\\
  \texttt{alicej@student.ethz.ch}
  \And Torsten Hoefler \\
  ETH Zurich\\
  Zurich 8092, Switzerland\\
  \texttt{htor@inf.ethz.ch}
}

\begin{document}

\maketitle

\begin{abstract}
With the recent success of embeddings in natural language processing, research has been conducted into applying similar methods to code analysis. Most works attempt to process the code directly or use a syntactic tree representation, treating it like sentences written in a natural language. However, none of the existing methods are sufficient to comprehend program semantics robustly, due to structural features such as function calls, branching, and interchangeable order of statements. In this paper, we propose a novel processing technique to learn code semantics, and apply it to a variety of program analysis tasks. In particular, we stipulate that a robust distributional hypothesis of code applies to both human- and machine-generated programs. Following this hypothesis, we define an embedding space, inst2vec, based on an Intermediate Representation (IR) of the code that is independent of the source programming language. We provide a novel definition of contextual flow for this IR, leveraging both the underlying data- and control-flow of the program. We then analyze the embeddings qualitatively using analogies and clustering, and evaluate the learned representation on three different high-level tasks. We show that even without fine-tuning, a single RNN architecture and fixed inst2vec embeddings outperform specialized approaches for performance prediction (compute device mapping, optimal thread coarsening); and algorithm classification from raw code (104 classes), where we set a new state-of-the-art. 
\end{abstract}

\section{Introduction}

The emergence of the ``Big Data era'' manifests in the form of a dramatic increase in accessible code. In the year 2017 alone, GitHub reports~\cite{github} approximately 1 billion git commits (code modification uploads) written in 337 different programming languages. Sifting through, categorizing, and understanding code thus becomes an essential task for a variety of fields. 
Applications include identifying code duplication, performance prediction, algorithm detection for alternative code suggestion (guided programming), vulnerability analysis, and malicious code detection.
These tasks are challenging, as code can be modified such that it syntactically differs (for instance, via different or reordered operations, or written in a different language altogether), but remains semantically equivalent (i.e., produces the same result). However, these tasks are also ideal for machine learning, since they can be represented as classic regression and classification problems.

In order to mechanize code comprehension, the research community typically employs reinforcement learning and stochastic compilation for \textit{super-optimization}~\cite{superopt,stoke}; or borrows concepts from Natural Language Processing (NLP) for human-authored code, relying on the following hypothesis:
\begin{quote}
	\textbf{The naturalness hypothesis~\cite{allamanis2017survey}.} \textit{Software is a form of human communication; software
		corpora have similar statistical properties to natural language corpora; and these
		properties can be exploited to build better software engineering tools.}
\end{quote}
For NLP-based approaches, input code is usually processed into tokens (e.g., keywords, braces)~\cite{cummins2017b} or other representations~\cite{allamanis2017:dfg,alon18code2vec,Park2012}, and optionally undergoes embedding in a continuous lower-dimensional space. In the spirit of the successful \texttt{word2vec} model~\cite{w2v_efficientestimation, w2v_distribrep}, the mapping to the embedding space is learned by pairing a token with its surrounding tokens. Following this process, RNNs~\cite{elman90} are trained on sequences of such tokens. This model has been successfully used for NLP-like tasks, such as summarization~\cite{allamanis16summarization}, function name prediction~\cite{alon18code2vec}, and algorithm classification~\cite{treecnn}. 

Although the results for stochastic code optimization and NLP embeddings are promising, two issues arise. Firstly, in prior works, the source programming language (or machine code for optimization) is fixed, which does not reflect the plethora of languages, nor generalizes to future languages. Secondly, existing methods process tokens (or instructions) sequentially, targeting function- and loop-free code. Such codes, however, do not represent the majority of the applications.

\begin{figure}[t]
	\centering
	\includegraphics[height=1.4in,trim={0cm 7cm 0cm 7cm},clip]{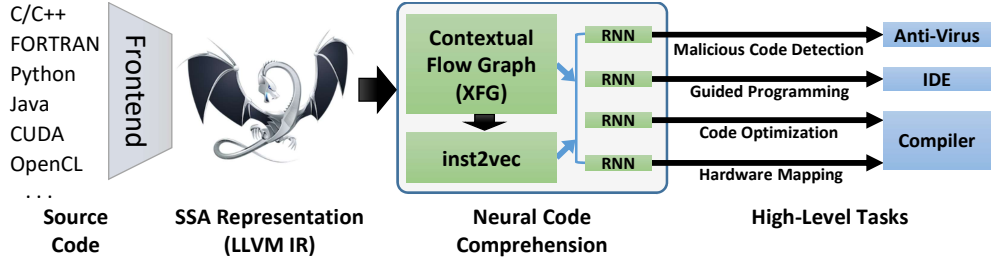}
	\caption{Component overview of the Neural Code Comprehension pipeline.}
	\vspace{-1em}
	\label{fig:overview}
\end{figure}

This paper presents Neural Code Comprehension\footnote{Code, datasets, trained embeddings, and results available at~~\url{https://www.github.com/spcl/ncc}}: a general-purpose processing pipeline geared towards representing code semantics in a robust and learnable manner. The pipeline, depicted in Fig. \ref{fig:overview}, accepts code in various source languages and converts it to statements in an Intermediate Representation (IR), using the LLVM Compiler Infrastructure~\cite{llvm}. The LLVM IR, which is explained in detail in Section \ref{sec:xfg}, is then processed to a robust representation that we call \textit{conteXtual Flow Graphs (XFGs)}. XFGs are constructed from both the data- and control-flow of the code, thus inherently supporting loops and function calls. In turn, the XFG structure is used to train an embedding space for individual statements, called \texttt{inst2vec} (from the word ``instruction''), which is fed to RNNs for a variety of high-level tasks. 

Neural Code Comprehension is evaluated on multiple levels, using clustering and analogies for \texttt{inst2vec}, as well as three different code comprehension tasks for XFGs: algorithm classification; heterogeneous compute device (e.g., CPU, GPU) mapping; and optimal thread coarsening factor prediction, which model the runtime of an application without running it.
Our datasets contain CPU and GPU code written in C, C++, OpenCL, and FORTRAN, though LLVM supports additional languages such as Python, Rust, Swift, Go, and CUDA.
Our work makes the following contributions:
\begin{itemize}
	\item We formulate a robust distributional hypothesis for code, from which we draw a novel distributed representation of code statements based on contextual flow and LLVM IR.
	\item We detail the construction of the XFG, \emph{the first representation designed specifically for statement embeddings that combines data and control flow}.
	\item We evaluate the representation using clustering, analogies, semantic tests, and three fundamentally different high-level code learning tasks.
	\item Using one simple LSTM architecture and fixed pre-trained embeddings, we match or surpass the best-performing approaches in each task, including specialized DNN architectures.
\end{itemize}

\section{Related Work}

Distributed representations of code were first suggested by Allamanis et al.~\cite{allamanis2015}, followed by several works leveraging embeddings to apply NLP techniques to programming languages~\cite{allamanis2017survey, Vechev:2016}. 

\paragraph{Code Representation} 
Previous research focuses on embedding high-level programming languages such as Java~\cite{Dam2016, Gu:2016}, C~\cite{Levy17}, or OpenCL~\cite{cummins2017b} in the form of \textit{tokens} or statements, as well as lower level representations such as object code~\cite{Levy17}. To the best of our knowledge, however, no attempt has been made to train embeddings for compiler IRs prior to this work. As for representing the context of a token, which is necessary for training embeddings, some works rely on lexicographical locality~\cite{allamanis2015,cummins2017b,Dam2016}, whereas others exploit the structural nature of code, using Data Flow Graphs~\cite{allamanis2017:dfg}, Control Flow Graphs \cite{Nobre:2016, Park2012, XU:2017}, Abstract Syntax Trees (ASTs)~\cite{bielik16, Gu:2016}, paths in the AST~\cite{Alon2018:GeneralPathBased}, or an augmented AST, for instance with additional edges connecting different uses and updates of syntax tokens corresponding to variables~\cite{allamanis:2017:graphs}. We differ from all previous approaches by introducing contextual flow, a graph representation that captures both data and control dependencies. In compiler research, similar graphs exist but have not been successfully exploited for machine learning. Examples include the Program Dependence Graph (PDG)~\cite{Ferrante:1987:PDG} and the IR known as Sea of Nodes~\cite{Click:1995b,Click:1995}. Unlike these representations, our graphs are not designed to be optimized by a compiler nor translated to machine code, which allows us to introduce ambiguity (e.g., ignoring parameter order) in favor of preserving context. Other works applying Machine Learning techniques to PDGs exist: Hsiao et al.~\cite{Hsiao:2014:pdg} use PDGs to compute n-gram models for program analysis, and Wang et al.~\cite{Wang:2015:pdg} use them for detecting copy direction among programs using Extreme Learning Machines. However, our work is the first to leverage a hybrid of control and data flow for the training of embeddings.

\paragraph{Automated Tasks on Code} 
Learned representations of code are commonly used for two types of tasks: uncovering program semantics or optimizing programs. For the former task, code embeddings have been used to perform function or variable naming~\cite{allamanis2015, alon18code2vec}, clone detection \cite{White:2016}, code completion~\cite{Raychev:2014,Yang:2017},
summarization~\cite{allamanis16summarization}, and algorithm classification~\cite{treecnn}. 
As for program optimization, research has been conducted on automatic feature generation for code~\cite{Leather:2009, Namolaru2010}; and Cummins et al.~\cite{cummins2017b} notably leverage embeddings of 
OpenCL code to predict optimal device mapping and thread coarsening factors. Their work differs from ours in that the method is restricted to the OpenCL language, and that they process programs in a sequential order, which does not capture complex code structures.
Furthermore, the state-of-the-art in automatic tuning for program optimization \cite{balaprakash13} uses surrogate performance models and active learning, and does not take code semantics into account.

\paragraph{Embedding Evaluation} Previous works that use code embeddings do not evaluate the quality of the trained space on its own merit, but rather through the performance of subsequent (downstream) tasks. One exception is Allamanis et al.~\cite{allamanis2015}, who present empirical evidence of vector similarities for similar method names.
To the best of our knowledge, we are the first to quantify the quality of a code embedding space itself in the form of clustering, syntactic analogies, semantic analogies, and categorical distance tests.

\section{A Robust Distributional Hypothesis of Code}
\label{sec:hypothesis}

The linguistic Distributional Hypothesis~\cite{harris81dh,pantel05} is given by: \emph{Words that occur in the same contexts tend to have similar meanings}. We stipulate that code, which describes a sequence of operations to a processor, behaves similarly, and paraphrase this hypothesis to:
\begin{center}
	\itshape \textbf{Statements} that occur in the same \textbf{contexts} tend to have \textbf{similar semantics}.
\end{center}
However, the above wording is vague, due to the possible meanings of the highlighted elements. 
Below we attempt to provide adequate definitions, upon which we build a learnable code representation.

\paragraph{Statements}
To choose the right abstraction for statements, we take two concerns into account: universality and uniformity. As stated above, source code comes in many languages and thus fixating on a single one would hinder universality. At the other extreme, machine code (assembly) is target-specific, containing specialized instructions and relying on hardware characteristics, such as registers and memory architectures. As for uniformity, in a high-level language one statement may represent simple arithmetics, multiple operations, or even class definitions (for example, the Java statement \texttt{button.setOnClickListener(new View.OnClickListener()\{...\})}). On the other hand, assembly is too limited, since instructions are reused for different purposes. We thus wish to choose statements that are independent of the source language, as well as the hardware architecture.

\paragraph{Context}
The definition of a context for code statements should also be carefully considered. 
We define context as \emph{statements whose execution directly depends on each other}. Learning from consecutive statements in code does not necessarily fulfill this definition, as, for example, a programmer may use a variable in the first line of a function, but only use it again in the last line. Moreover, such long-term relationships may vanish when using RNNs and attention learning. It is possible to determine the data dependencies of each statement by analyzing dataflow, however, branches and function calls do not necessarily generate such dependencies. Another way of representing execution dependence is through the notion of causality (i.e., the ``happens-before'' relation~\cite{lamport78}), which can be used to complement dataflow.
In our representation, context is the union of data dependence and execution dependence, thereby capturing both relations.

\paragraph{Similarity}
To define similarity, one first needs to define the \textit{semantics} of a statement. We draw the definition of semantics from Operational Semantics in programming language theory, which refers to the effects (e.g., preconditions, postconditions) of each computational step in a given program. In this paper, we specifically assume that each statement modifies the system state in a certain way (e.g., adds two numbers) and consumes resources (e.g., uses registers and floating-point units). It follows that semantic similarity can be defined by two statements consuming the same resources or modifying the system state in a similar way. Using this definition, two versions of the same algorithm with different variable types would be synonymous.

\section{Contextual Flow Processing}
\label{sec:xfg}

\begin{figure}
	\centering
	\begin{minipage}[b]{0.35\textwidth}
		\centering
		\begin{subfigure}[t]{\linewidth}
			\includegraphics[page=1,width=0.5\linewidth,clip,trim={0cm 13.9cm 26cm 0.25cm}]{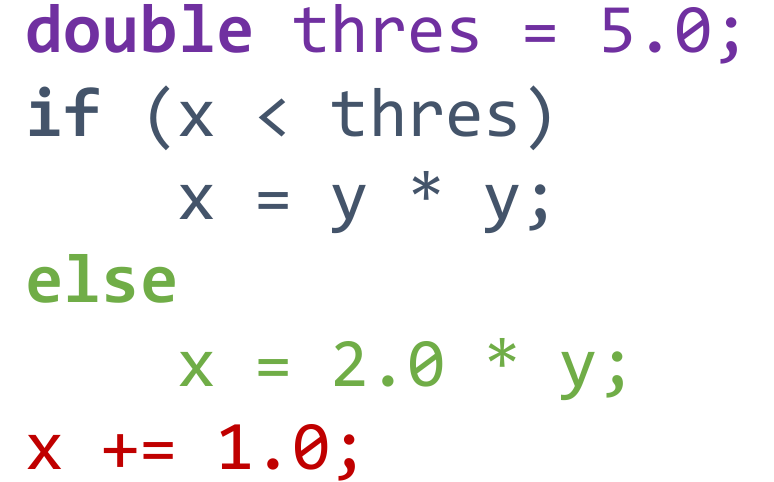}
			\caption{Source code}		
			\label{fig:xfg:source}
		\end{subfigure}\vspace{0.1in}
		\begin{subfigure}[t]{\linewidth}
			\includegraphics[page=3,width=\linewidth,clip,trim={0cm 11.3cm 18cm 0cm}]{figures/xfg-paged}	
			\caption{LLVM IR}
			\label{fig:xfg:ir}
		\end{subfigure}
	\end{minipage}
	\begin{subfigure}[t]{0.3\textwidth}
		\centering
		\includegraphics[page=4,height=1.9in,clip,trim={0cm 11.3cm 27.3cm 0cm}]{figures/xfg-paged}
		\caption{Dataflow basic blocks}
		\label{fig:xfg:basicblocks}
	\end{subfigure}
	\begin{subfigure}[t]{0.3\textwidth}
		\centering
		\includegraphics[page=5,height=1.9in,clip,trim={0cm 11.35cm 27.25cm 0cm}]{figures/xfg-paged}
		\caption{Contextual Flow Graph}
		\label{fig:xfg:xfg}
	\end{subfigure}
	\caption{Contextual flow processing scheme.}
	\label{fig:xfg}
	\vspace{-1.5em}
\end{figure}

The aforementioned statements and contexts cannot be directly extracted from source code, but rather require processing akin to partial compilation (e.g., dataflow extraction). In this section, we briefly describe a popular compilation pipeline and proposed modifications to create a learnable vocabulary of statements and their context.

\subsection{Compilation, Static Single Assignment, and LLVM IR}

\begin{figure}[h!]
	\centering
	\vspace{-0.5em}
	\begin{minipage}[b]{0.6\textwidth}
		\centering
		\includegraphics[height=0.4in,clip,trim={0.1cm 1.4cm 1.05cm 0.5cm}]{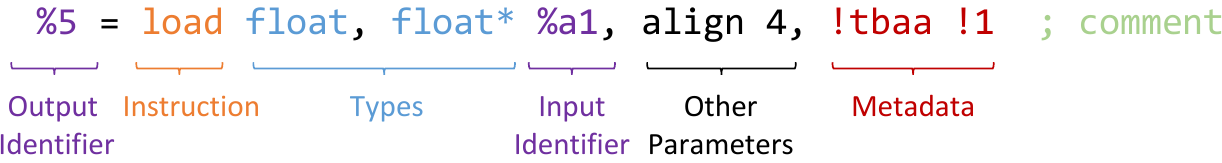}
		\caption{Anatomy of an LLVM IR statement.}
		\label{fig:llvmir}		
	\end{minipage}
	\begin{minipage}[b]{0.39\textwidth}
		\scriptsize
		\centering
		\verb|x0 = a * b;   |\\
		\verb|x1 = c * c;   |\\
		\verb|x2 = x0 + x1; |\\
		\verb|x3 = x + x2;  |\\
		\caption{SSA of \texttt{x += (a*b)+(c*c)}.}
		\label{fig:ssa}
	\end{minipage}
	\vspace{-1.5em}
\end{figure}

Major contemporary compilers, such as GCC and LLVM, support multiple programming languages and hardware targets. To avoid duplication in code optimization techniques, they enforce a strict separation between the source language (frontend), an Intermediate Representation (IR) that can be optimized, and the target machine code (backend) that should be mapped to a specific hardware. In particular, the LLVM IR~\cite{llvmlang} supports various architectures (e.g., GPUs), and can represent optimized code (e.g., using vector registers) inherently. 
Figures \ref{fig:xfg:source} and \ref{fig:xfg:ir} depict an example code and its LLVM IR equivalent, and the structure of an LLVM IR statement is shown in Fig. \ref{fig:llvmir}.

In the LLVM infrastructure, the IR is given in Static Single Assignment (SSA) form~\cite{ssa}. Briefly, an SSA IR ensures that every variable is assigned only once, which makes it easy to track dataflow between IR statements, as shown in Fig. \ref{fig:ssa}.
To overcome analysis issues resulting from control-flow, such as loops, SSA defines $\phi$-expressions. These expressions enumerate all possible outcomes that can lead to a variable (depending on the runtime control-flow), and can be used to optimize code across branches. In Fig. \ref{fig:xfg:ir}, the identifier \texttt{\%4} is constructed from a $\phi$-expression that can take either the value of \texttt{\%2} or \texttt{\%3}, depending on the value of \texttt{x}.

\subsection{Contextual Flow Graphs}

To analyze dataflow for optimization, LLVM divides the IR statements into ``basic blocks'', which contain no control-flow divergence, illustrated in Fig. \ref{fig:xfg:basicblocks}. Within a basic block, statements naturally create traceable dataflow as SSA lists data dependencies in the form of input identifiers (even if conditional), and assigns the results to a single identifier. However, as shown in Section \ref{sec:hypothesis}, dataflow alone does not suffice to provide context for a given statement, e.g., when in the vicinity of a branch. 
Therefore, we define a representation that incorporates both the relative data- and control-flow of a statement, which we call the \textbf{conteXtual Flow Graph (XFG)}.

XFGs (e.g., Fig. \ref{fig:xfg:xfg}) are directed multigraphs, where two nodes can be connected by more than one edge. XFG nodes can either be variables or label identifiers (e.g., basic block, function name), appearing in the figure as ovals or rectangles respectively. Correspondingly, an edge either represents data-dependence (in black), carrying an LLVM IR statement; or execution dependence (light blue).

\textbf{XFG Construction}~~~We generate XFGs incrementally from LLVM IR, as follows:
\begin{enumerate}
	\item \vspace{-0.5em}Read LLVM IR statements once, storing function names and return statements.
	\item Second pass over the statements, adding nodes and edges according to the following rule-set:
	\begin{enumerate}
		\item Data dependencies within a basic block are connected.
		\item Inter-block dependencies (e.g., $\phi$-expressions) are both connected directly and through the label identifier (statement-less edges).
		\item Identifiers without a dataflow parent are connected to their root (label or program root).
	\end{enumerate}
\end{enumerate}

It follows that XFGs create paths through dataflow as well as branches, loops, and functions (including recursion).
Owing to the two passes, as well as the linear-time construction of LLVM IR \cite{sreedhar95}, XFGs are constructed in $\bigO{n}$ for a program with $n$ SSA statements. This is especially valuable when learning over large code corpora, such as Tensorflow.

\textbf{External Code}~~~Calls to external code (e.g., libraries, frameworks) can be divided into two categories: statically- and dynamically-linked. If the code is accessible during compilation (header-only frameworks and static libraries), LLVM IR is available and the statements are traversed as part of the XFG. In the dynamic case, the library code is not included and is represented as a \texttt{call} statement.

\section{\texttt{inst2vec}: Embedding Statements in Continuous Space}
\label{sec:inst2vec}

With XFGs providing a notion of context, we can now train an embedding space for individual statements. To support learnability, desiderata for such a space include: (a) statements that are in close proximity should have similar artifacts on a system (i.e., use the same resources); and (b) changing the same attributes (e.g., data type) for different instructions should result in a similar offset in the space. We train LLVM IR statement embeddings using the skip-gram model~\cite{w2v_distribrep}, following preprocessing to limit the vocabulary size.

\subsection{Statement Preprocessing and Training}

\paragraph{Preprocessing}
First, we filter out comments and metadata from statements. Then, identifiers and immediate values (numeric constants, strings) are replaced with \texttt{\%ID} and \texttt{<INT/FLOAT/STRING>} respectively, where immediate values are fed separately to downstream RNNs. Lastly, data structures are ``inlined'', that is, their contents are encoded within the statement. Fig. \ref{fig:inst2vecpp} lists statements before and after preprocessing.
\begin{figure}[h]
	\vspace{-0.5em}
	\begin{subfigure}{.45\textwidth}
		\centering
		\scriptsize
\begin{verbatim}
store float %250, float* %82, align 4, !tbaa !1
%10 = fadd fast float %9, 1.3
%8 = load %"struct.aaa"*, %"struct.aaa"** %2
\end{verbatim}
		\vspace{-1em}
		\caption{LLVM IR}
		\label{fig:inst2vecpp:llvm}
	\end{subfigure}
	\quad
	\begin{subfigure}{.5\textwidth}
		\centering
		\scriptsize
\begin{verbatim}
store float %ID, float* %ID, align 4
%ID = fadd fast float %ID, <FLOAT>
%ID = load { float, float }*, { float, float }** %ID
\end{verbatim}
		\vspace{-1em}
		\caption{\texttt{inst2vec} statements}
		\label{fig:inst2vecpp:i2v}
	\end{subfigure}
	\vspace{-0.5em}
	\caption{Before and after preprocessing LLVM IR to \texttt{inst2vec} statements.}
	\label{fig:inst2vecpp}
\end{figure}

\textbf{Dataset}~~~Table \ref{tbl:i2vstats} summarizes the code corpora and vocabulary statistics of the \texttt{inst2vec} dataset.
We choose corpora from different disciplines, including high-performance computing, benchmarks, operating systems, climate sciences, computer vision, machine learning (using Tensorflow's own source code), and synthetically-generated programs. The code in the dataset is written in C, C++, FORTRAN, and OpenCL, and is compiled for Intel CPUs as well as NVIDIA and AMD GPUs. 
The files in the dataset were compiled to LLVM IR with Clang~\cite{clang} and Flang~\cite{flang}, using compilation flags from the original code (if available) and randomly chosen compiler optimization (e.g., \texttt{-ffast-math}) and target architecture flags.

For the synthetic corpus, we use both C code and the Eigen~\cite{eigen} C++ library. In particular, random linear algebra operations are procedurally generated from high-level templates using different parameters, such as data types, operations, and dimensions.

 \begin{table}[t]
 	\caption{\texttt{inst2vec} training dataset statistics}
 	\label{tbl:i2vstats}
 	\centering
 	\scriptsize
 	\begin{tabular}{llrrrr}
 		\toprule
 		Discipline & Dataset & Files & LLVM IR & Vocabulary & XFG Stmt. \\
 		&         &       & Lines   & Size       & Pairs\\
 		\midrule
 		Machine Learning & Tensorflow~\cite{tensorflow2015-whitepaper} & 2,492 & 16,943,893 & 220,554 &  260,250,973 \\\addlinespace
 		High-Performance Computing & AMD APP SDK~\cite{amd} & 123 & 1,304,669 & 4,146 &  45,081,359\\
 		& BLAS~\cite{blas}       & 300 & 280,782 & 566 &  283,856 \\\addlinespace     
 		Benchmarks       & NAS~\cite{snu-npb} & 268 & 572,521 & 1,793 &   1,701,968  \\
 		& Parboil~\cite{parboil} & 151 & 118,575 & 2,175 &  151,916 \\
 		& PolybenchGPU~\cite{polybench}& 40 & 33,601 & 577 &  40,975 \\
 		& Rodinia~\cite{rodinia} & 92 & 103,296 & 3,861 &  266,354 \\
 		& SHOC~\cite{shoc} & 112 & 399,287 & 3,381 &  12,096,508 \\\addlinespace
 		Scientific Computing & COSMO~\cite{cosmo}     & 161 & 152,127 & 2,344 & 2,338,153\\\addlinespace					     
 		Operating Systems& Linux kernel~\cite{linuxkernel} & 1,988 & 2,544,245 & 136,545 &  5,271,179 \\\addlinespace
 		
 		Computer Vision  & OpenCV~\cite{opencv} & 442 & 1,908,683 & 39,920 &  10,313,451 \\
 		& NVIDIA samples~\cite{nvidiasamples} & 60 & 43,563 & 2,467 &  74,915 \\\addlinespace
 		Synthetic		& Synthetic & 17,801 & 26,045,547 & 113,763 &  303,054,685 \\\addlinespace
 		
 		\midrule
 		Total (Combined) & --- & 24,030 & 50,450,789 & 8,565 &  640,926,292 \\
 		\bottomrule
 	\end{tabular}
 	\vspace{-1em}
 \end{table}

\paragraph{Setup and Training}

Given a set of XFGs created from the LLVM IR files, we generate neighboring statement pairs up to a certain context size, following the skip-gram model~\cite{w2v_distribrep}. A context of size $N$ includes all statement pairs that are connected by a path shorter or equal to $N$. 
To obtain the pairs, we construct a dual graph in which statements are nodes, omitting duplicate edges. 
Following this process, we discard statements that occur less than 300 times in the dataset, pairs of identical statements, and perform subsampling of frequent pairs, similarly to Mikolov et al.~\cite{w2v_distribrep}.
We train \texttt{inst2vec} with an embedding dimension of 200 for 5 epochs using Tensorflow~\cite{tensorflow2015-whitepaper}. The Adam optimizer~\cite{adam} is used with the default published hyperparameters and softmax cross-entropy loss.

\subsection{Evaluation}

\paragraph{Clustering}

\begin{figure}[t]
	\centering
	\begin{subfigure}[b]{0.3\textwidth}
		\centering
		\includegraphics[width=\linewidth,clip,trim={0cm 0cm 5cm 0cm}]{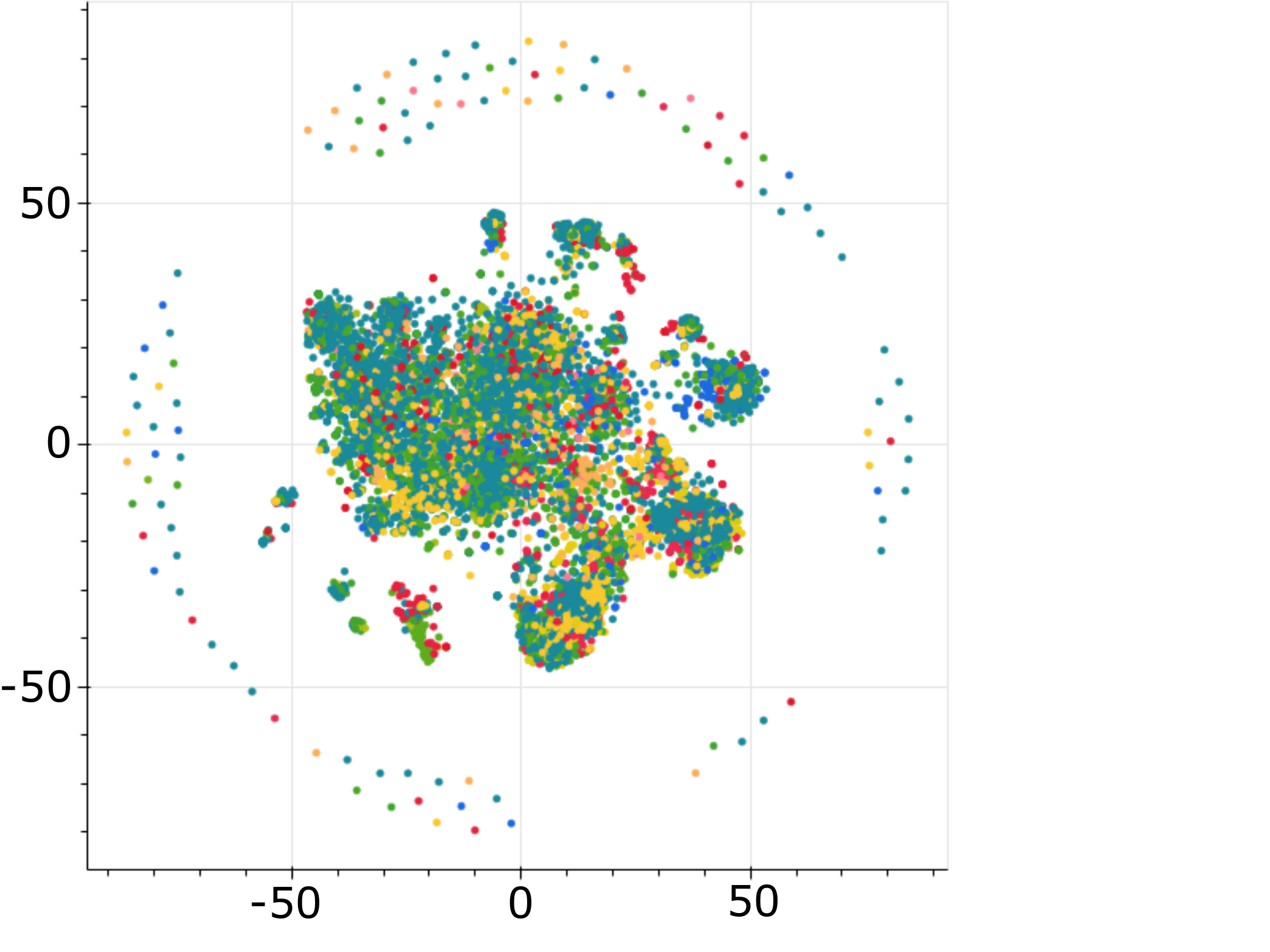}
		\caption{Context size = 1}
		\label{fig:tsne:cw1}
	\end{subfigure}
	\quad
	\begin{subfigure}[b]{0.3\textwidth}
		\centering
		\includegraphics[width=\linewidth,clip,trim={0cm 0cm 5cm 0cm}]{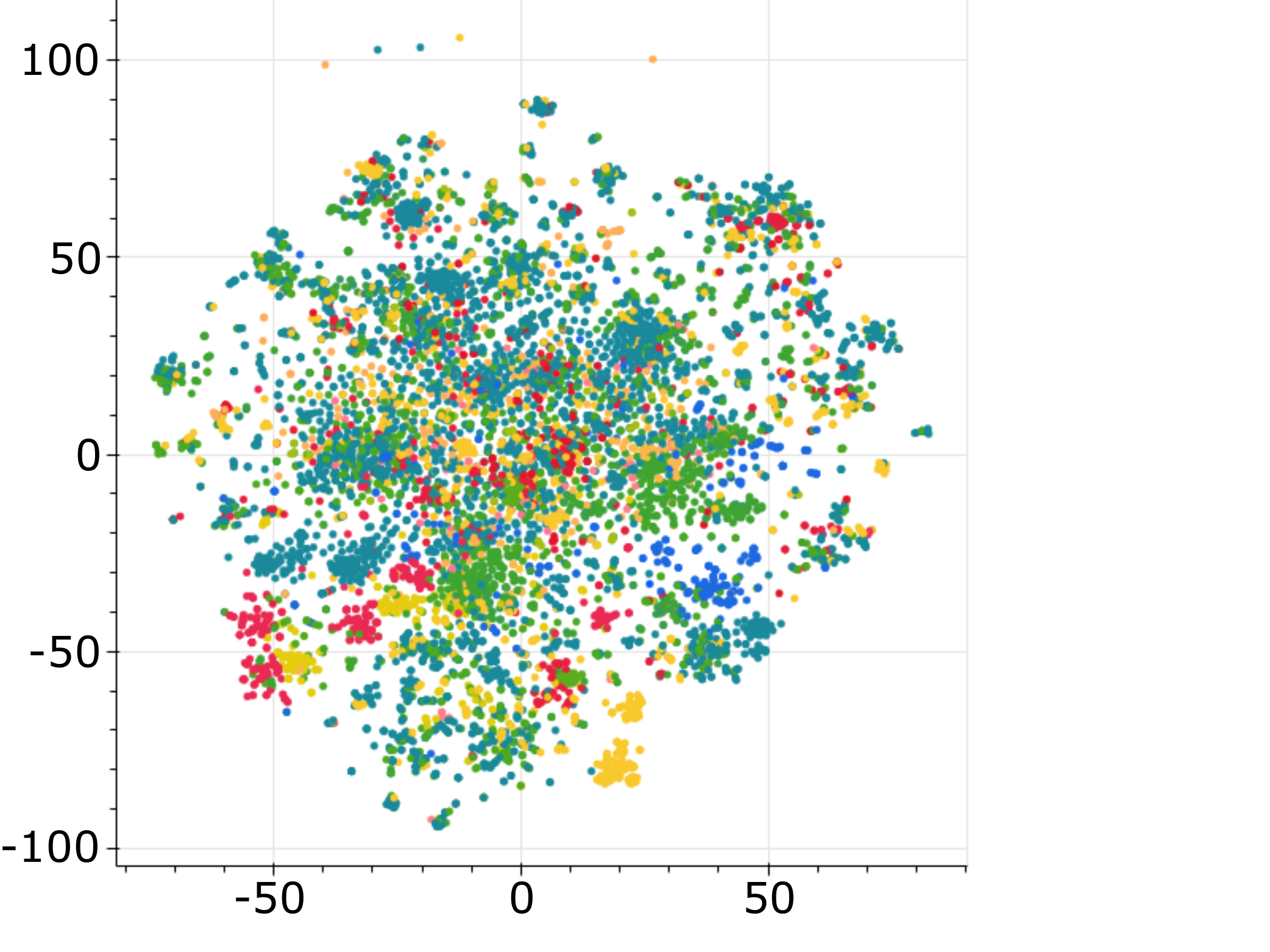}
		\caption{Context size = 2}
		\label{fig:tsne:cw2}
	\end{subfigure}
	\quad
	\begin{subfigure}[b]{0.3\textwidth}
		\centering
		\includegraphics[width=\linewidth,clip,trim={0cm -0.1cm 4.8cm 0cm}]{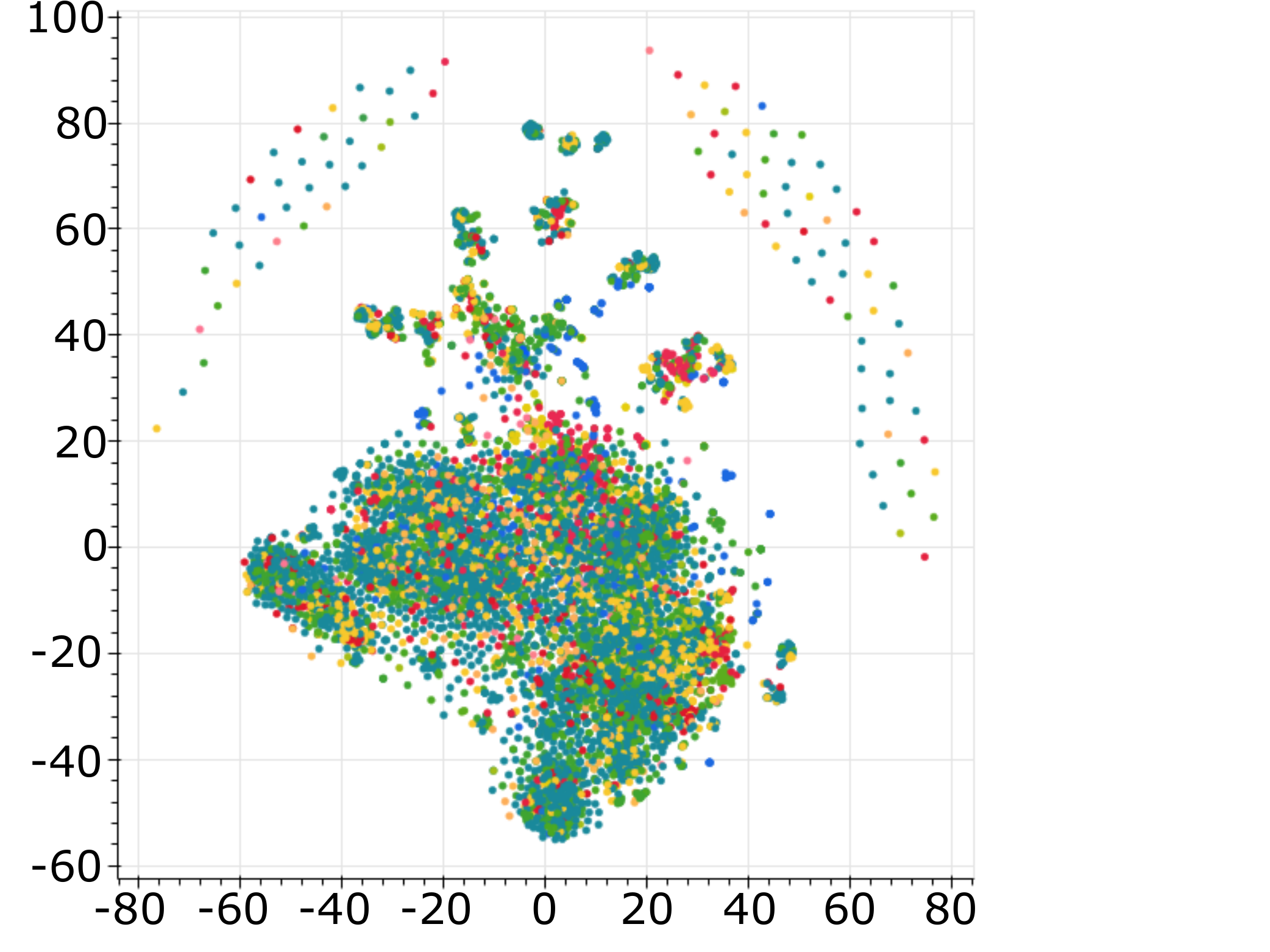}
		\caption{Context size = 3}
		\label{fig:tsne:cw3}
	\end{subfigure}
	\caption{Two-dimensional t-SNE plots for learned embeddings (best viewed in color).}
	\label{fig:tsne}
\end{figure}

Fig. \ref{fig:tsne} depicts the t-SNE~\cite{tsne} plots for trained \texttt{inst2vec} spaces with different XFG context sizes, colored by statement and data type (legend in  Appendix \ref{app:clustering}). In the plots, we see that both a context size of 1 statement in each direction (Fig. \ref{fig:tsne:cw1}) or 3 statements (Fig. \ref{fig:tsne:cw3}) generate large, multi-type clusters, as well as outliers. This phenomenon eventually contributes to a lower final analogy score, due to inappropriate representation of inter-statement relations, as can be seen below. Owing to these results, we choose a context size of 2 statements (Fig. \ref{fig:tsne:cw2}), which mostly consists of separate, monochromatic clusters, indicating strong clustering w.r.t. instruction and data types. While data type syntactic clusters are unsurprising, their existence is not trivial, since the dataset contains diverse codebases rather than copies of the same functions with different types.

An example of a \textit{semantically}-similar statement cluster can be found in data structures. In particular, the top-5 nearest neighbors of operations on the complex data type ``\texttt{std::complex<float>}'' include ``\texttt{2 x float}'' (i.e., a vector type). In fact, LLVM IR represents the complex data type as \texttt{\{float, float\}}, so this property is generalized to any user-defined data structure (\texttt{struct}) with two floats.

\paragraph{Analogies and Tests}

We also evaluate \texttt{inst2vec} by automatically generating a list of statement analogies (``a'' is to ``b'' as ``c'' is to ``?'', or ``\texttt{a:b; c:?}'') that appear in our vocabulary using the LLVM IR syntax. We then use the embeddings to find the result by computing \texttt{a-b+c} and asking whether the result is in the top-5 neighbors (cosine distance). Additionally, we automatically create relative distance expressions using the LLVM IR reference categories~\cite{llvmlang} of the form $d(a,b)<d(a,c)$ to test whether statements that use different resources are further away than those who use the same.

Table~\ref{tbl:analogies} shows the analogy and test results for \texttt{inst2vec} trained on XFG as well as on CFG (control flow-only) and DFG (data flow-only) for different context sizes. The analogies are divided into different categories, including data types (i.e., transitions between types), options (e.g., fast math), conversions (e.g., bit casting, extension, truncation), and data structures (e.g., vector-type equivalents of structures). Below are examples of a type analogy:\\
{\scriptsize
		\verb|        %ID = add i64 %ID, %ID   :   %ID = fadd float %ID, %ID;|\\
		\verb|        %ID = sub i64 %ID, %ID   :?  %ID = fsub float %ID, %ID|}\\
and a data structure analogy:\\ {\scriptsize
		\verb|        %ID = extractvalue { double, double } %ID, 0   :   %ID = extractelement <2 x double> %ID, <TYP> 0;|\\
		\verb|        %ID = extractvalue { double, double } %ID, 1   :?  %ID = extractelement <2 x double> %ID, <TYP> 1|
	}

The results confirm that over all scores, a context size of 2 is the best-performing configuration, and show that the XFG representation is more complete and leads to better embeddings than taking into account control or data flow alone.

\begin{table}[h]
	\caption{Analogy and test scores for \texttt{inst2vec}}
	\label{tbl:analogies}
	\centering
	\scriptsize
	\begin{tabular}{llrrrrr}
		\toprule
		\multirow{2}{1cm}{Context type} & \multirow{2}{1cm}{Context Size} & \multicolumn{2}{c}{Syntactic Analogies} & \multicolumn{2}{c}{Semantic Analogies} & Semantic Distance Test  \\
        \cmidrule(r){3-4} \cmidrule(r){5-6}
		& & Types & Options & Conversions & Data Structures & \\
		\midrule
		
		CFG
		& 1& 0     (0\thinspace\%) &      1  (1.89\thinspace\%) &       1 (0.07\thinspace\%) &     0     (0\thinspace\%)& 51.59\thinspace\%\\
		& 2& 1  (0.18\thinspace\%) &      1  (1.89\thinspace\%) &       0    (0\thinspace\%) &     0     (0\thinspace\%)& 50.47\thinspace\%\\
		& 3& 0     (0\thinspace\%) &      1  (1.89\thinspace\%) &       4 (0.27\thinspace\%) &     0     (0\thinspace\%)& 53.79\thinspace\%\\
		\midrule
		
		DFG
		& 1& 53  (9.46\thinspace\%) &     12 (22.64\thinspace\%) &       2 (0.13\thinspace\%) &     4 (50.00\thinspace\%)& 56.79\thinspace\%\\
		& 2& 71 (12.68\thinspace\%) &     12 (22.64\thinspace\%) &      12 (0.80\thinspace\%) &     3 (37.50\thinspace\%)& 57.44\thinspace\%\\
		& 3& 67 (22.32\thinspace\%) &     18 (33.96\thinspace\%) &      40 (2.65\thinspace\%) &     4 (50.00\thinspace\%)& 60.38\thinspace\%\\
		\midrule
		
		XFG
		& 1&     101 (18.04\thinspace\%) &     13 (24.53\thinspace\%) &     100 (6.63\thinspace\%) &     3 (37.50\thinspace\%)& 60.98\thinspace\%\\
		& 2& \bf 226 (40.36\thinspace\%) & \bf 45 (84.91\thinspace\%) & \bf 134 (8.89\thinspace\%) & \bf 7 (87.50\thinspace\%)&\bf 79.12\thinspace\%\\
		& 3&     125 (22.32\thinspace\%) &     24 (45.28\thinspace\%) &      48 (3.18\thinspace\%) & \bf 7 (87.50\thinspace\%)& 62.56\thinspace\%\\
		\bottomrule
	\end{tabular}
\end{table}

\section{Code Comprehension Experiments}

In this section, we evaluate \texttt{inst2vec} on three different tasks, comparing with manually-extracted features and state-of-the-art specialized deep learning approaches. Throughout all tasks, we use \emph{the same neural network architecture} and our pre-trained embedding matrix from Section \ref{sec:inst2vec}, which remains fixed during training.

\textbf{Training}~~~Our recurrent network (see schematic description in the Appendix~\ref{app:ncc-architecture}) consists of an \texttt{inst2vec} input with an XFG context size of 2, followed by two stacked LSTM \cite{lstm} layers with 200 units in each layer, batch normalization~\cite{bn}, a dense 32-neuron layer with ReLU activations, and output units matching the number of classes. The loss function is a categorical cross-entropy trained using Adam~\cite{adam} with the default hyperparameters. Additionally, for the compute device mapping and optimal thread coarsening factor prediction tasks, we train the LLVM IR statements with the immediate values that were stripped from them during preprocessing (see Section \ref{sec:inst2vec}). Further details are given in Appendix~\ref{app:method:immval}.

\textbf{Datasets}~~~The algorithm classification task uses the POJ-104~\cite{treecnn} dataset\footnote{\url{https://sites.google.com/site/treebasedcnn/}}, collected from a Pedagogical Open Judge system. The dataset contains 104 program classes written by 500 different people (randomly selected subset per class). For the compute device mapping and optimal thread coarsening factor prediction tasks, we use an OpenCL code dataset\footnote{\url{https://www.github.com/ChrisCummins/paper-end2end-dl}} provided by Cummins et al.~\cite{cummins2017b}.

\subsection{Algorithm Classification} \label{sec:tasks:classifyalg}

Using \texttt{inst2vec}, we construct an RNN that reads embedded source code and outputs a predicted program class. We compare our approach with Tree-Based CNNs (TBCNN)~\cite{treecnn}, the best-performing algorithm classifier in the POJ-104 dataset. TBCNN constructs embeddings from Astract Syntax Tree nodes of source code, and employs two specialized layers: tree convolutions and dynamic pooling. Their network comprises 5 layers, where convolution and fully connected layers are 600-dimensional. Our data preparation follows the experiment conducted by Mou et al.~\cite{treecnn}, splitting the dataset 3:1:1 for training, validation, and testing. To compile the programs successfully, we prepend \texttt{\#include} statements to each file. Data augmentation is then applied on the training set by compiling each file 8 times with different flags (\texttt{-O\{0-3\}}, \texttt{-ffast-math}). 

\begin{table}[h]
	\caption{Algorithm classification test accuracy}
	\label{res:task:classifyalg}
	\centering
	\scriptsize
	\begin{tabular}{lcccc}
		\toprule
		Metric        &  Surface Features~\cite{treecnn} & RNN~\cite{treecnn} & TBCNN~\cite{treecnn}  & inst2vec \\
		& (RBF SVM + Bag-of-Trees) & & & \\
		\midrule
		Test Accuracy [\%] & 88.2 & 84.8 & 94.0 & \textbf{94.83} \\
		\bottomrule
	\end{tabular}
\end{table}

Table \ref{res:task:classifyalg} compares \texttt{inst2vec} (trained for 100 epochs) with the reported results of Mou et al.~\cite{treecnn}, which contain TBCNN as well as a 600-cell RNN and a manual feature extraction approach (Surface Features). The results show that \texttt{inst2vec} sets a new state-of-the-art with a 13.8\thinspace\% decrease in error, even though the dataset used to generate the embeddings \emph{does not include POJ-104} (see Table \ref{tbl:i2vstats}).

\subsection{Heterogeneous Compute Device Mapping}
\label{sec:tasks:devmap}
Next, we use Neural Code Comprehension to predict whether a given OpenCL program will run
faster on a CPU (Intel Core i7-3820) or a GPU (AMD Tahiti 7970 and NVIDIA GTX 970) given its code, input data size, and \textit{work-group size} (i.e., number of threads that work in a group with shared memory). To achieve that, we use the same experimental methodology presented by Cummins et al.~\cite{cummins2017b}, removing their specialized OpenCL source rewriter and replacing their code token embeddings with our XFGs and \texttt{inst2vec}. We concatenate the data and work-group sizes to the network inputs, and train with stratified 10-fold cross-validation. We repeat the training $10$ times with random initialization of the network's weights and report the best result.

In Table \ref{res:task:devmap}, \texttt{inst2vec} and \texttt{inst2vec-imm} (i.e., with immediate value handling) are compared with a manual code feature extraction approach by Grewe et al.~\cite{devmap-grewe} and DeepTune~\cite{cummins2017b}, in terms of runtime prediction accuracies and resulting speedup. The baseline for the speedup is a static mapping, which selects the device that yields the best average case performance over all programs in the data set: in the case of AMD Tahiti versus Intel i7-3820, that is the CPU and in the case of NVIDIA GTX versus Intel i7-3820, it is the GPU. The results indicate that \texttt{inst2vec} outperforms Grewe et al.~and is on-par with DeepTune. We believe that the better predictions in DeepTune are the result of training the embedding matrix in tandem with the high-level task, thereby specializing it to the dataset. This specialized training is, however, surpassed by taking immediate values into account during training. We present the result of the best immediate value handling method in Table~\ref{res:task:devmap} (\texttt{inst2vec-imm}), and the exhaustive results can be found in Appendix~\ref{app:res:immval}.

\begin{table}
  \caption{Heterogeneous device mapping results}
  \label{res:task:devmap}
  \centering
  \scriptsize
  \begin{tabular}{lccccc}
	\toprule
	Architecture        & \multicolumn{5}{c}{Prediction Accuracy [\%]} \\ 
	\cmidrule(r){2-6}  
	& GPU & Grewe et al.~\cite{devmap-grewe} & DeepTune~\cite{cummins2017b}  & inst2vec & inst2vec-imm  \\
	\midrule
	AMD Tahiti 7970     & 41.18 & 73.38 & 83.68 & 82.79 & \textbf{88.09} \\
	NVIDIA GTX 970      & 56.91 & 72.94 & 80.29 & 82.06 & \textbf{86.62} \\
	
	\midrule
	&\multicolumn{5}{c}{Speedup}\\
	\cmidrule(r){2-6}
	& GPU & Grewe et al. & DeepTune  & inst2vec & inst2vec-imm \\
	\midrule
	AMD Tahiti 7970     & 3.26 & 2.91 & 3.34 & 3.42 & \textbf{3.47}\\
	NVIDIA GTX 970      & 1.00 & 1.26 & 1.41 & 1.42 & \textbf{1.44}\\
	
	\bottomrule
  \end{tabular}

\end{table}

\subsection{Optimal Thread Coarsening Factor Prediction}

Our third example predicts the best-performing \textit{thread coarsening factor}, a measure of the amount of work done per GPU thread, for a given OpenCL code. We again compare the achieved speedups of \texttt{inst2vec} with manual features~\cite{threadcoars-magni}, DeepTune, and DeepTune with transfer learning applied from the task in Section \ref{sec:tasks:devmap} (denoted by DeepTune-TL). Possible values for the coarsening factor are $1$ (baseline for speedups), $2$, $4$, $8$, $16$, and $32$. The results in Table \ref{res:task:cf:compare_speedups} show that while \texttt{inst2vec} yields better speedups than DeepTune-TL in only half of the cases (possibly due to the embedding specialization in DeepTune), the manually-extracted features are consistently outperformed by \texttt{inst2vec}. Moreover, \texttt{inst2vec-imm} is consistently on-par with DeepTune, but improves inconsistently on \texttt{inst2vec} (on the AMD Tahiti and the NVIDIA GTX only), and fails to outperform DeepTune-TL. This can be explained by the small size of the training data for this task ($17$ programs with $6$ different thread coarsening factors for each hardware platform). The optimal device mapping task (Section ~\ref{sec:tasks:devmap}), on the other hand, features $680$ programs for each platform.

\begin{table}[h]
  \caption{Speedups achieved by coarsening threads}
  \label{res:task:cf:compare_speedups}
  \centering
  \scriptsize
  \begin{tabular}{lccccc}
    \toprule
    Computing Platform
                       & Magni et al.~\cite{threadcoars-magni} & DeepTune~\cite{cummins2017b}  & DeepTune-TL~\cite{cummins2017b} & inst2vec & inst2vec-imm \\
    \midrule
    AMD Radeon HD 5900  & 1.21         & 1.10      & 1.17        & \textbf{1.37} & 1.28 \\
    AMD Tahiti 7970     & 1.01         & 1.05      & \textbf{1.23}        & 1.10 & 1.18 \\
    NVIDIA GTX 480      & 0.86         & 1.10      & \textbf{1.14}        & 1.07 & 1.11 \\
    NVIDIA Tesla K20c   & 0.94         & 0.99      & 0.93        & \textbf{1.06} & 1.00 \\
    \bottomrule
  \end{tabular}
\end{table}

\section{Conclusion}

In this paper, we have empirically shown that semantics of statements can be successfully recovered from their context alone. This recovery relies both on proper granularity, where we propose to use filtered LLVM IR instructions; and on the grouping of statements, for which we use a mixture of data- and control-flow. We use our proposed representation to perform three high-level classification and prediction tasks, outperforming all manually-extracted features and achieving results that are on-par with (and better than) two inherently different state-of-the-art specialized DNN solutions.

With this work, we attempt to pave the way towards mechanized code comprehension via machine learning, whether the code was authored by a human or automatically-generated. Further research could be conducted in various directions. Rather than directly using statements, the representation may be refined using part-based models, which have already been applied successfully in language models~\cite{santos14charwnn}. 
\texttt{inst2vec} can also be used as a basis for neural code interpretation, using a modified Differentiable Neural Computer~\cite{dnc} to enable execution of arbitrary code over DNNs.

\subsubsection*{Acknowledgments}
We wish to thank Theodoros Theodoridis, Kfir Levy, Tobias Grosser, and Yunyan Guo for fruitful discussions. The authors also acknowledge MeteoSwiss, and thank Hussein Harake, Colin McMurtrie, and the whole CSCS team for granting access to the Greina machines, and for their excellent technical support. TBN is supported by the ETH Postdoctoral Fellowship and Marie Curie Actions for People COFUND program.

% References
\bibliographystyle{plain}
\bibliography{references}

\begin{thebibliography}{10}

\bibitem{tensorflow2015-whitepaper}
Mart\'{\i}n Abadi, Ashish Agarwal, Paul Barham, Eugene Brevdo, Zhifeng Chen,
  Craig Citro, Greg~S. Corrado, Andy Davis, Jeffrey Dean, Matthieu Devin,
  Sanjay Ghemawat, Ian Goodfellow, Andrew Harp, Geoffrey Irving, Michael Isard,
  Yangqing Jia, Rafal Jozefowicz, Lukasz Kaiser, Manjunath Kudlur, Josh
  Levenberg, Dandelion Man\'{e}, Rajat Monga, Sherry Moore, Derek Murray, Chris
  Olah, Mike Schuster, Jonathon Shlens, Benoit Steiner, Ilya Sutskever, Kunal
  Talwar, Paul Tucker, Vincent Vanhoucke, Vijay Vasudevan, Fernanda Vi\'{e}gas,
  Oriol Vinyals, Pete Warden, Martin Wattenberg, Martin Wicke, Yuan Yu, and
  Xiaoqiang Zheng.
\newblock {TensorFlow}: Large-scale machine learning on heterogeneous systems,
  2015.
\newblock Software available from tensorflow.org.

\bibitem{allamanis2015}
Miltiadis Allamanis, Earl~T. Barr, Christian Bird, and Charles Sutton.
\newblock Suggesting accurate method and class names.
\newblock In {\em Proceedings of the 2015 10th Joint Meeting on Foundations of
  Software Engineering}, ESEC/FSE 2015, pages 38--49, New York, NY, USA, 2015.
  ACM.

\bibitem{allamanis2017survey}
Miltiadis Allamanis, Earl~T. Barr, Premkumar~T. Devanbu, and Charles~A. Sutton.
\newblock A survey of machine learning for big code and naturalness.
\newblock {\em CoRR}, abs/1709.06182, 2017.

\bibitem{allamanis2017:dfg}
Miltiadis Allamanis and Marc Brockschmidt.
\newblock Smartpaste: Learning to adapt source code.
\newblock {\em CoRR}, abs/1705.07867, 2017.

\bibitem{allamanis:2017:graphs}
Miltiadis Allamanis, Marc Brockschmidt, and Mahmoud Khademi.
\newblock Learning to represent programs with graphs.
\newblock {\em CoRR}, abs/1711.00740, 2017.

\bibitem{allamanis16summarization}
Miltiadis Allamanis, Hao Peng, and Charles~A. Sutton.
\newblock A convolutional attention network for extreme summarization of source
  code.
\newblock {\em CoRR}, abs/1602.03001, 2016.

\bibitem{alon18code2vec}
Uri Alon, Meital Zilberstein, Omer Levy, and Eran Yahav.
\newblock code2vec: Learning distributed representations of code.
\newblock {\em CoRR}, abs/1803.09473, 2018.

\bibitem{Alon2018:GeneralPathBased}
Uri Alon, Meital Zilberstein, Omer Levy, and Eran Yahav.
\newblock A general path-based representation for predicting program
  properties.
\newblock {\em CoRR}, abs/1803.09544, 2018.

\bibitem{amd}
{AMD}.
\newblock {AMD OpenCL} accelerated parallel processing {SDK}.
\newblock
  \url{https://developer.amd.com/amd-accelerated-parallel-processing-app-sdk/}.

\bibitem{balaprakash13}
P.~Balaprakash, R.~B. Gramacy, and S.~M. Wild.
\newblock Active-learning-based surrogate models for empirical performance
  tuning.
\newblock In {\em 2013 IEEE International Conference on Cluster Computing
  (CLUSTER)}, pages 1--8, Sept 2013.

\bibitem{cosmo}
M.~{Baldauf}, A.~{Seifert}, J.~{F{\"o}rstner}, D.~{Majewski},
  M.~{Raschendorfer}, and T.~{Reinhardt}.
\newblock {Operational Convective-Scale Numerical Weather Prediction with the
  {COSMO} Model: Description and Sensitivities}.
\newblock {\em Monthly Weather Review}, 139:3887--3905, December 2011.

\bibitem{bielik16}
Pavol Bielik, Veselin Raychev, and Martin Vechev.
\newblock {PHOG}: Probabilistic model for code.
\newblock In Maria~Florina Balcan and Kilian~Q. Weinberger, editors, {\em
  Proceedings of The 33rd International Conference on Machine Learning},
  volume~48 of {\em Proceedings of Machine Learning Research}, pages
  2933--2942, New York, New York, USA, June 2016. PMLR.

\bibitem{superopt}
Rudy Bunel, Alban Desmaison, M.~Pawan Kumar, Philip H.~S. Torr, and Pushmeet
  Kohli.
\newblock Learning to superoptimize programs.
\newblock {\em International Conference on Learning Representations}, 2017.

\bibitem{rodinia}
Shuai Che, Michael Boyer, Jiayuan Meng, David Tarjan, Jeremy~W. Sheaffer,
  Sang-Ha Lee, and Kevin Skadron.
\newblock Rodinia: A benchmark suite for heterogeneous computing.
\newblock In {\em Proceedings of the 2009 IEEE International Symposium on
  Workload Characterization (IISWC)}, IISWC '09, pages 44--54, Washington, DC,
  USA, 2009. IEEE Computer Society.

\bibitem{Click:1995b}
Cliff Click and Keith~D. Cooper.
\newblock Combining analyses, combining optimizations.
\newblock {\em ACM Transactions on Programming Languages and Systems}, 17,
  1995.

\bibitem{Click:1995}
Cliff Click and Michael Paleczny.
\newblock A simple graph-based intermediate representation.
\newblock {\em SIGPLAN Not.}, 30(3):35--49, March 1995.

\bibitem{nvidiasamples}
NVIDIA Corporation.
\newblock {CUDA}.
\newblock \url{http://developer.nvidia.com/object/cuda.html}.

\bibitem{cummins2017b}
Chris Cummins, Pavlos Petoumenos, Zheng Wang, and Hugh Leather.
\newblock End-to-end deep learning of optimization heuristics.
\newblock In {\em PACT}. ACM, 2017.

\bibitem{ssa}
Ron Cytron, Jeanne Ferrante, Barry~K. Rosen, Mark~N. Wegman, and F.~Kenneth
  Zadeck.
\newblock Efficiently computing static single assignment form and the control
  dependence graph.
\newblock {\em ACM Trans. Program. Lang. Syst.}, 13(4):451--490, October 1991.

\bibitem{Dam2016}
Hoa~Khanh Dam, Truyen Tran, and Trang Pham.
\newblock A deep language model for software code.
\newblock {\em CoRR}, abs/1608.02715, 2016.

\bibitem{shoc}
Anthony Danalis, Gabriel Marin, Collin McCurdy, Jeremy S.~Meredith, Philip
  Roth, Kyle Spafford, Vinod Tipparaju, and Jeffrey Vetter.
\newblock The {Scalable HeterOgeneous Computing (SHOC)} benchmark suite.
\newblock pages 63--74, January 2010.

\bibitem{blas}
Jack Dongarra.
\newblock Basic linear algebra subprograms technical forum standard.
\newblock page 1 — 111, 2002.

\bibitem{elman90}
Jeffrey~L. Elman.
\newblock Finding structure in time.
\newblock {\em Cognitive Science}, 14(2):179 -- 211, 1990.

\bibitem{Ferrante:1987:PDG}
Jeanne Ferrante, Karl~J. Ottenstein, and Joe~D. Warren.
\newblock The program dependence graph and its use in optimization.
\newblock {\em ACM Trans. Program. Lang. Syst.}, 9(3):319--349, July 1987.

\bibitem{github}
GitHub.
\newblock {GitHub Octoverse}.
\newblock \url{https://octoverse.github.com/}, 2017.

\bibitem{Glorot:2011}
Xavier Glorot, Antoine Bordes, and Yoshua Bengio.
\newblock Deep sparse rectifier neural networks.
\newblock 2011.

\bibitem{polybench}
Scott Grauer-Gray, Lifan Xu, Robert Searles, Sudhee Ayalasomayajula, and John
  Cavazos.
\newblock Auto-tuning a high-level language targeted to {GPU} codes.
\newblock 2012.

\bibitem{dnc}
Alex Graves, Greg Wayne, Malcolm Reynolds, Tim Harley, Ivo Danihelka, Agnieszka
  Grabska-Barwi{\'n}ska, Sergio~G{\'o}mez Colmenarejo, Edward Grefenstette,
  Tiago Ramalho, John Agapiou, et~al.
\newblock Hybrid computing using a neural network with dynamic external memory.
\newblock {\em Nature}, 538(7626):471--476, 2016.

\bibitem{devmap-grewe}
Dominik Grewe, Zheng Wang, and Michael O'Boyle.
\newblock Portable mapping of data parallel programs to {OpenCL} for
  heterogeneous systems.
\newblock pages 1--10, February 2013.

\bibitem{Gu:2016}
Xiaodong Gu, Hongyu Zhang, Dongmei Zhang, and Sunghun Kim.
\newblock Deep {API} learning.
\newblock In {\em Proceedings of the 2016 24th ACM SIGSOFT International
  Symposium on Foundations of Software Engineering}, FSE 2016, pages 631--642,
  New York, NY, USA, 2016. ACM.

\bibitem{eigen}
Ga\"{e}l Guennebaud and Beno\^{i}t~Jacob et~al.
\newblock Eigen v3.
\newblock \url{http://eigen.tuxfamily.org}, 2010.

\bibitem{harris81dh}
Zellig~S. Harris.
\newblock {\em Distributional Structure}, pages 3--22.
\newblock Springer Netherlands, Dordrecht, 1981.

\bibitem{lstm}
Sepp Hochreiter and J\"{u}rgen Schmidhuber.
\newblock Long short-term memory.
\newblock {\em Neural Computation}, 9(8):1735--1780, 1997.

\bibitem{Hsiao:2014:pdg}
Chun-Hung Hsiao, Michael Cafarella, and Satish Narayanasamy.
\newblock Using web corpus statistics for program analysis.
\newblock In {\em Proceedings of the 2014 ACM International Conference on
  Object Oriented Programming Systems Languages \& Applications}, OOPSLA '14,
  pages 49--65, New York, NY, USA, 2014. ACM.

\bibitem{bn}
Sergey Ioffe and Christian Szegedy.
\newblock Batch normalization: Accelerating deep network training by reducing
  internal covariate shift.
\newblock In {\em Proceedings of the 32Nd International Conference on
  International Conference on Machine Learning - Volume 37}, ICML'15, pages
  448--456. JMLR.org, 2015.

\bibitem{opencv}
Itseez.
\newblock Open source computer vision library.
\newblock \url{https://github.com/itseez/opencv}, 2015.

\bibitem{adam}
Diederik~P. Kingma and Jimmy Ba.
\newblock Adam: {A} method for stochastic optimization.
\newblock {\em CoRR}, abs/1412.6980, 2014.

\bibitem{lamport78}
Leslie Lamport.
\newblock Time, clocks, and the ordering of events in a distributed system.
\newblock {\em Commun. ACM}, 21(7):558--565, July 1978.

\bibitem{llvm}
Chris Lattner and Vikram Adve.
\newblock {LLVM}: a compilation framework for lifelong program analysis
  transformation.
\newblock In {\em International Symposium on Code Generation and Optimization,
  2004. CGO 2004.}, pages 75--86, March 2004.

\bibitem{Leather:2009}
Hugh Leather, Edwin Bonilla, and Michael O'Boyle.
\newblock Automatic feature generation for machine learning based optimizing
  compilation.
\newblock In {\em Proceedings of the 7th Annual IEEE/ACM International
  Symposium on Code Generation and Optimization}, CGO '09, pages 81--91,
  Washington, DC, USA, 2009. IEEE Computer Society.

\bibitem{Levy17}
Dor Levy and Lior Wolf.
\newblock Learning to align the source code to the compiled object code.
\newblock In {\em Proceedings of the 34th International Conference on Machine
  Learning, {ICML} 2017, Sydney, NSW, Australia, 6-11 August 2017}, pages
  2043--2051, 2017.

\bibitem{linuxkernel}
Linux.
\newblock Linux kernel source code (version 4.15.1).
\newblock \url{https://www.kernel.org/}.

\bibitem{flang}
{LLVM}.
\newblock Flang: a {FORTRAN} compiler frontend for {LLVM}.
\newblock \url{https://github.com/flang-compiler/flang}.

\bibitem{clang}
{LLVM}.
\newblock Clang: a {C} language family frontend for {LLVM} v4.0.0.
\newblock \url{http://clang.llvm.org/}, 2017.

\bibitem{llvmlang}
LLVM.
\newblock {LLVM} language reference manual.
\newblock \url{https://llvm.org/docs/LangRef.html}, 2018.

\bibitem{threadcoars-magni}
Alberto Magni, Christophe Dubach, and Michael O'Boyle.
\newblock Automatic optimization of thread-coarsening for graphics processors.
\newblock In {\em Proceedings of the 23rd International Conference on Parallel
  Architectures and Compilation}, PACT '14, pages 455--466, New York, NY, USA,
  2014. ACM.

\bibitem{w2v_efficientestimation}
Tomas Mikolov, Kai Chen, Greg Corrado, and Jeffrey Dean.
\newblock Efficient estimation of word representations in vector space.
\newblock {\em CoRR}, abs/1301.3781, 2013.

\bibitem{w2v_distribrep}
Tomas Mikolov, Ilya Sutskever, Kai Chen, Greg Corrado, and Jeffrey Dean.
\newblock Distributed representations of words and phrases and their
  compositionality.
\newblock In {\em Proceedings of the 26th International Conference on Neural
  Information Processing Systems - Volume 2}, NIPS'13, pages 3111--3119, USA,
  2013. Curran Associates Inc.

\bibitem{treecnn}
Lili Mou, Ge~Li, Lu~Zhang, Tao Wang, and Zhi Jin.
\newblock Convolutional neural networks over tree structures for programming
  language processing.
\newblock In {\em Proceedings of the Thirtieth AAAI Conference on Artificial
  Intelligence}, AAAI'16, pages 1287--1293. AAAI Press, 2016.

\bibitem{Namolaru2010}
Mircea Namolaru, Albert Cohen, Grigori Fursin, Ayal Zaks, and Ari Freund.
\newblock Practical aggregation of semantical program properties for machine
  learning based optimization.
\newblock In {\em Proceedings of the 2010 International Conference on
  Compilers, Architectures and Synthesis for Embedded Systems}, CASES '10,
  pages 197--206, New York, NY, USA, 2010. ACM.

\bibitem{Nobre:2016}
Ricardo Nobre, Luiz G.~A. Martins, and Jo\~{a}o M.~P. Cardoso.
\newblock A graph-based iterative compiler pass selection and phase ordering
  approach.
\newblock In {\em Proceedings of the 17th ACM SIGPLAN/SIGBED Conference on
  Languages, Compilers, Tools, and Theory for Embedded Systems}, LCTES 2016,
  pages 21--30, New York, NY, USA, 2016. ACM.

\bibitem{pantel05}
Patrick Pantel.
\newblock Inducing ontological co-occurrence vectors.
\newblock In {\em Proceedings of the 43rd Annual Meeting on Association for
  Computational Linguistics}, ACL '05, pages 125--132, Stroudsburg, PA, USA,
  2005. Association for Computational Linguistics.

\bibitem{Park2012}
Eunjung Park, John Cavazos, and Marco~A. Alvarez.
\newblock Using graph-based program characterization for predictive modeling.
\newblock In {\em Proceedings of the Tenth International Symposium on Code
  Generation and Optimization}, CGO '12, pages 196--206, New York, NY, USA,
  2012. ACM.

\bibitem{Raychev:2014}
Veselin Raychev, Martin Vechev, and Eran Yahav.
\newblock Code completion with statistical language models.
\newblock {\em SIGPLAN Not.}, 49(6):419--428, June 2014.

\bibitem{santos14charwnn}
Cicero~Dos Santos and Bianca Zadrozny.
\newblock Learning character-level representations for part-of-speech tagging.
\newblock In Eric~P. Xing and Tony Jebara, editors, {\em Proceedings of the
  31st International Conference on Machine Learning}, volume~32 of {\em
  Proceedings of Machine Learning Research}, pages 1818--1826, Bejing, China,
  June 2014. PMLR.

\bibitem{stoke}
Eric Schkufza, Rahul Sharma, and Alex Aiken.
\newblock Stochastic superoptimization.
\newblock In {\em Proceedings of the Eighteenth International Conference on
  Architectural Support for Programming Languages and Operating Systems},
  ASPLOS '13, pages 305--316, New York, NY, USA, 2013. ACM.

\bibitem{snu-npb}
Sangmin Seo, Gangwon Jo, and Jaejin Lee.
\newblock Performance characterization of the nas parallel benchmarks in
  opencl.
\newblock In {\em Proceedings of the 2011 IEEE International Symposium on
  Workload Characterization}, IISWC '11, pages 137--148, Washington, DC, USA,
  2011. IEEE Computer Society.

\bibitem{sreedhar95}
Vugranam~C. Sreedhar and Guang~R. Gao.
\newblock A linear time algorithm for placing phi-nodes.
\newblock In {\em Proceedings of the 22Nd ACM SIGPLAN-SIGACT Symposium on
  Principles of Programming Languages}, POPL '95, pages 62--73, New York, NY,
  USA, 1995. ACM.

\bibitem{parboil}
John~A. Stratton, Christopher Rodrigues, I-Jui Sung, Nady Obeid, Li-Wen Chang,
  Nasser Anssari, Geng~Daniel Liu, and Wen-mei~W. Hwu.
\newblock Parboil: A revised benchmark suite for scientific and commercial
  throughput computing.
\newblock {\em Center for Reliable and High-Performance Computing}, 2012.

\bibitem{tsne}
Laurens van~der Maaten and Geoffrey Hinton.
\newblock Visualizing data using {t-SNE}.
\newblock {\em Journal of Machine Learning Research}, 9:2579--2605, 2008.

\bibitem{Vechev:2016}
Martin Vechev and Eran Yahav.
\newblock Programming with "big code".
\newblock {\em Found. Trends Program. Lang.}, 3(4):231--284, December 2016.

\bibitem{Wang:2015:pdg}
Baoezeng Wang, Xiaochun Yang, and Guoren Wang.
\newblock Detecting copy directions among programs using extreme learning
  machines.
\newblock 2015:1--15, 05 2015.

\bibitem{White:2016}
Martin White, Michele Tufano, Christopher Vendome, and Denys Poshyvanyk.
\newblock Deep learning code fragments for code clone detection.
\newblock In {\em Proceedings of the 31st IEEE/ACM International Conference on
  Automated Software Engineering}, ASE 2016, pages 87--98, New York, NY, USA,
  2016. ACM.

\bibitem{XU:2017}
Xiaojun Xu, Chang Liu, Qian Feng, Heng Yin, Le~Song, and Dawn Song.
\newblock Neural network-based graph embedding for cross-platform binary code
  similarity detection.
\newblock {\em CoRR}, abs/1708.06525, 2017.

\bibitem{Yang:2017}
Yixiao Yang, Yu~Jiang, Ming Gu, Jiaguang Sun, Jian Gao, and Han Liu.
\newblock A language model for statements of software code.
\newblock In {\em Proceedings of the 32Nd IEEE/ACM International Conference on
  Automated Software Engineering}, ASE 2017, pages 682--687, Piscataway, NJ,
  USA, 2017. IEEE Press.

\end{thebibliography}

\clearpage
\appendix
\section{Statement Categories for \texttt{inst2vec} Clustering Results}
\label{app:clustering}

Table \ref{tbl:i2v:clustering:color_legend} presents the mapping from colors to statement categories that appear in Fig. \ref{fig:tsne}. The following rules apply to the categories in the table:
\begin{enumerate}
	\item A \texttt{type operation} generally refers to an operation, a function call, or the definition of a function, that returns an instance of \texttt{type}.
	\item \texttt{type*} refers to a pointer of \texttt{type}. Asterisks could be chained for pointers-to-pointers.
	\item \texttt{<d x type>} is a vector of \texttt{d} elements of \texttt{type}.
	\item \texttt{[d x type]} is an array of \texttt{d} elements of \texttt{type}.
	\item \texttt{struct/class} denotes an aggregate structure (e.g., C \texttt{struct}) of multiple types, e.g., \texttt{\{type\_1, type\_2, ..., type\_n\}} in LLVM IR.
	\item \texttt{floating point} can refer to either single- or double-precision floating point values.
	\item \texttt{int} can refer to an integer of any bit-width.
	\item \texttt{void} categories (\texttt{call void}, \texttt{invoke void}) refer to calls/invocations of functions that have no return value.
	\item \texttt{conversion operations} denote type conversions within LLVM, which do not necessarily translate into code.
	\item \texttt{load function pointer}, \texttt{store function pointer} refer to instructions that read or write function pointers into memory, respectively.
\end{enumerate}

\begin{table}[h!]
	\caption{Statement category by color (Fig. \ref{fig:tsne} legend)}
	\label{tbl:i2v:clustering:color_legend}
	\centering
	\scriptsize
	\renewcommand*{\arraystretch}{1.3}
	\begin{tabular}{c l l}
		\toprule
		Color & Statement Category & Example \\
		\midrule
		
		\tikzcircle[fill={rgb,255:red,8; green,92; blue,248}]{4pt} & \texttt{<d x int>* operation} & \texttt{<\%ID> = load <2 x i64>*, <2 x i64>** <\%ID>, align 8}\\
		
		\tikzcircle[fill={rgb,255:red,28; green,105; blue,225}]{4pt} & \texttt{<d x int> operation} & \texttt{<\%ID> = and <8 x i32> <\%ID>, <\%ID>}\\
		
		\tikzcircle[fill={rgb,255:red,27; green,117; blue,202}]{4pt} & \texttt{<d x struct/class*> operation} & \tiny\texttt{store <2 x \{ i64, i64 \}*> <\%ID>, <2 x \{ i64, i64 \}*>* <\%ID>, align 8}\\
		
		\tikzcircle[fill={rgb,255:red,19; green,128; blue,179}]{4pt} & \texttt{struct/class* operation} & \tiny\texttt{<\%ID> = phi \{ float, float \}* [ <\%ID>, <\%ID> ], [ <\%ID>, <\%ID> ]}\\
		
		\tikzcircle[fill={rgb,255:red,25; green,137; blue,155}]{4pt} & \texttt{struct/class operation} & \texttt{<\%ID> = alloca \{ i32, i32 \}, align 4}\\
		
		\tikzcircle[fill={rgb,255:red,56; green,151; blue,103}]{4pt} & \texttt{int** operation} & \texttt{<\%ID> = phi i8** [ <\%ID>, <\%ID> ], [ <\%ID>, <\%ID> ]}\\
		
		\tikzcircle[fill={rgb,255:red,59; green,158; blue,74}]{4pt} & \texttt{int* operation} & \texttt{<\%ID> = load i8*, i8** <\%ID>, align 8}\\
		
		\tikzcircle[fill={rgb,255:red,63; green,164; blue,49}]{4pt} & \texttt{int operation} & \texttt{<\%ID> = add i16 <\%ID>, <INT>}\\

		\tikzcircle[fill={rgb,255:red,75; green,169; blue,34}]{4pt} & \texttt{type conversion operation} & \texttt{<\%ID> = bitcast <4 x i32> <\%ID> to <16 x i8>}\\
		
		\tikzcircle[fill={rgb,255:red,92; green,173; blue,30}]{4pt} & \texttt{global variable definition} & \texttt{<@ID> = global i32 <INT>, align 4}\\
		
		\tikzcircle[fill={rgb,255:red,123; green,180; blue,27}]{4pt} & \texttt{<d x int*> operation} & \texttt{<\%ID> = phi <4 x i8*> [ <\%ID>, <\%ID> ], [ <\%ID>, <\%ID> ]}\\
		
		\tikzcircle[fill={rgb,255:red,164; green,190; blue,23}]{4pt} & \texttt{load function pointer} & \tiny\texttt{<\%ID> = load \{ i32 (...)** \}*, \{ i32 (...)** \}** <\%ID>, align 8}\\
		
		\tikzcircle[fill={rgb,255:red,177; green,193; blue,21}]{4pt} & \texttt{store function pointer} & \texttt{store void ()* <@ID>, void ()** <\%ID>, align 8}\\
		
		\tikzcircle[fill={rgb,255:red,203; green,199; blue,17}]{4pt} & \texttt{floating point** operation} & \texttt{<\%ID> = phi float** [ <\%ID>, <\%ID> ], [ <\%ID>, <\%ID> ]}\\
		
		\tikzcircle[fill={rgb,255:red,216; green,202; blue,14}]{4pt} & \texttt{floating point* operation} & \texttt{<\%ID> = icmp eq double* <\%ID>, null}\\
		
		\tikzcircle[fill={rgb,255:red,229; green,205; blue,14}]{4pt} & \texttt{floating point operation} & \texttt{<\%ID> = getelementptr double, double* <\%ID>, i64 <\%ID>}\\
		
		\tikzcircle[fill={rgb,255:red,248; green,200; blue,44}]{4pt} & \texttt{call void} & \texttt{tail call void <@ID>(i64 <INT>)}\\
		
		\tikzcircle[fill={rgb,255:red,252; green,183; blue,76}]{4pt} & \texttt{other/misc.} & \texttt{cleanup};~~~ \texttt{unreachable}\\
		
		\tikzcircle[fill={rgb,255:red,253; green,174; blue,89}]{4pt} & \texttt{[d x [d x type]] operation} & \tiny\texttt{<\%ID> = getelementptr inbounds [8 x [256 x i32]], [8 x [256 x i32]]*}  \\
		
		\tikzcircle[fill={rgb,255:red,254; green,165; blue,100}]{4pt} & \texttt{[d x struct/class] operation} & \texttt{<\%ID> = alloca [5 x \{ i8*, i64 \}], align 8}\\
		
		\tikzcircle[fill={rgb,255:red,253; green,127; blue,139}]{4pt} & \texttt{[d x int] operation} & \texttt{<\%ID> = alloca [100 x i8], align 16}\\
		
		\tikzcircle[fill={rgb,255:red,253; green,117; blue,147}]{4pt} & \texttt{[d x floating point] operation} & \tiny\texttt{<\%ID> = getelementptr inbounds [1024 x double], [1024 x double]*}\\
		
		\tikzcircle[fill={rgb,255:red,238; green,48; blue,102}]{4pt} & \texttt{<d x floating point>* operation} & \texttt{<\%ID> = alloca <8 x float>*, align 8}\\
		
		\tikzcircle[fill={rgb,255:red,233; green,41; blue,82}]{4pt} & \texttt{<d x floating point> operation} & \texttt{<\%ID> = call <4 x float> <@ID>(float* <\%ID>)}\\
		
		\tikzcircle[fill={rgb,255:red,229; green,32; blue,62}]{4pt} & \texttt{void function definition} & \texttt{define linkonce\_odr void <@ID>(\{ i32 (...)** \}*) unnamed\_addr}\\
		
		\tikzcircle[fill={rgb,255:red,223; green,23; blue,42}]{4pt} & \texttt{invoke void} & \texttt{invoke void <@ID>(i8* <\%ID>) to label <\%ID> unwind label <\%ID>}\\
		
		\bottomrule
	\end{tabular}
	\renewcommand*{\arraystretch}{1}
\end{table}

\newpage
\section{Neural Code Comprehension: Network Architecture}
\label{app:ncc-architecture}

\begin{figure}[h]
	\centering
	\includegraphics[page=1,width=\textwidth,trim={0cm 6.5cm 3.5cm 0cm},clip]{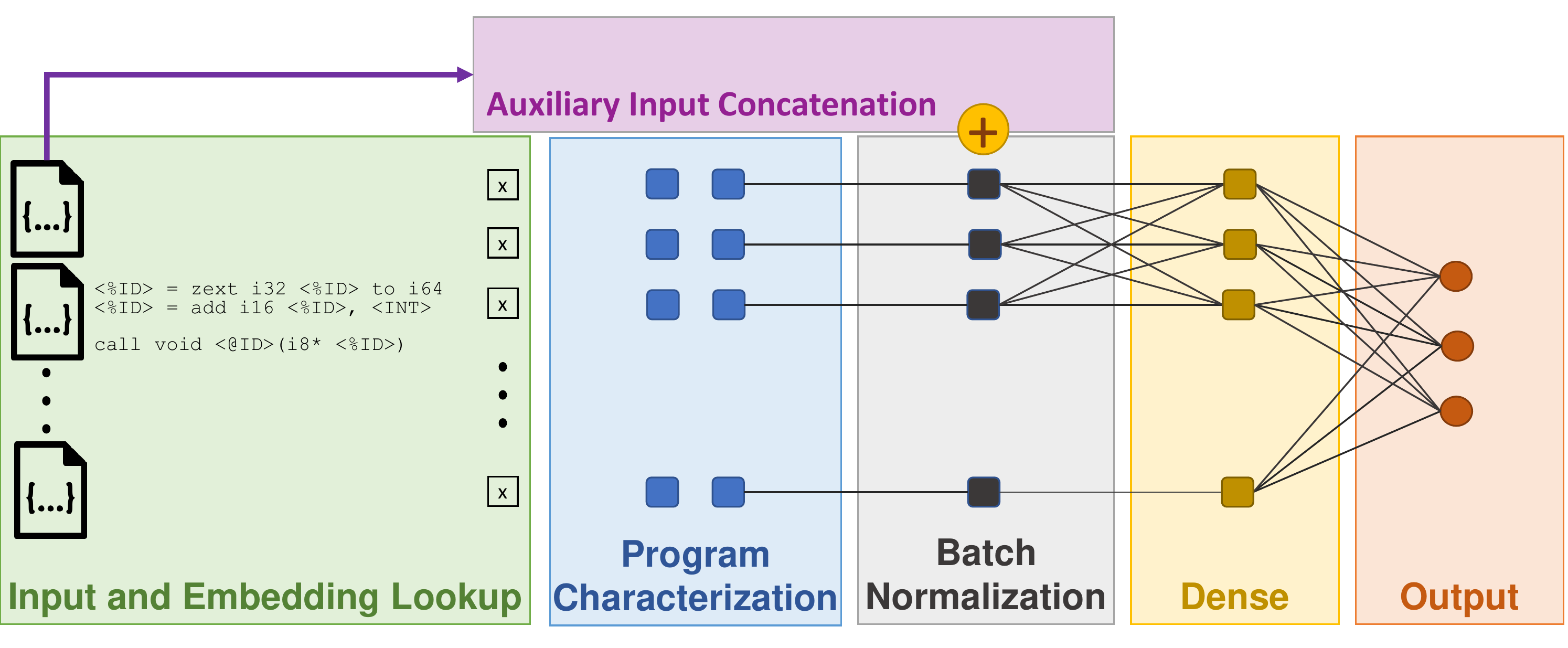}
	\caption{Neural Code Comprehension Network Architecture}
	\label{fig:ncc:arch}
\end{figure}

Fig. \ref{fig:ncc:arch} depicts the neural network architecture used for the high-level tasks in this paper. Below we describe each of the underlying layers in the network.

\textbf{Input and Embedding Lookup}~~~As an input, the Neural Code Comprehension architecture accepts programs as sequences of LLVM IR statements. Each statement is represented through its corresponding embedding vector, and for statements that are not in the \texttt{inst2vec} vocabulary, they are assigned the embedding vector corresponding to a predefined ``unknown'' token. The embedding layer remains fixed throughout the training of the code comprehension tasks (effectively, it acts as a simple lookup matrix), and no fine-tuning is applied to the vector representations.  

\textbf{Program Characterization}~~~The sequence of statement embedding vectors is passed to two layers of Long Short-Term Memory (LSTM)~\cite{lstm} cells. 
This program characterization layer transforms an input sequence of arbitrary length into a fixed-length vector that captures the properties of the processed program.	

\textbf{Auxiliary Input Concatenation (optional)}~~~Additional data may optionally be concatenated with the output of the two-layer LSTM at this point. This allows information that is only available at runtime (e.g., hardware parameters or data size) to be taken into account in the predictive modeling. 

\textbf{Batch normalization}~\cite{bn} is performed, and then the vector output of program characterization goes through a $32$-unit fully connected \textbf{dense} layer with rectifier (ReLU) activations~\cite{Glorot:2011}.  Finally, the \textbf{output} layer is another fully-connected layer, which features a number of units equal to the number of possible output categories. The output is given by a sigmoid activation function (output between $0$ and $1$), where the largest activation corresponds to the model's prediction.

\section{Training NCC with Immediate Values: Method Description}
\label{app:method:immval}

\begin{figure}
	\centering

	\begin{subfigure}[t]{\textwidth}
		\centering
		\includegraphics[page=2,height=1.3in,trim={0cm 6.5cm 3cm 0cm},clip]{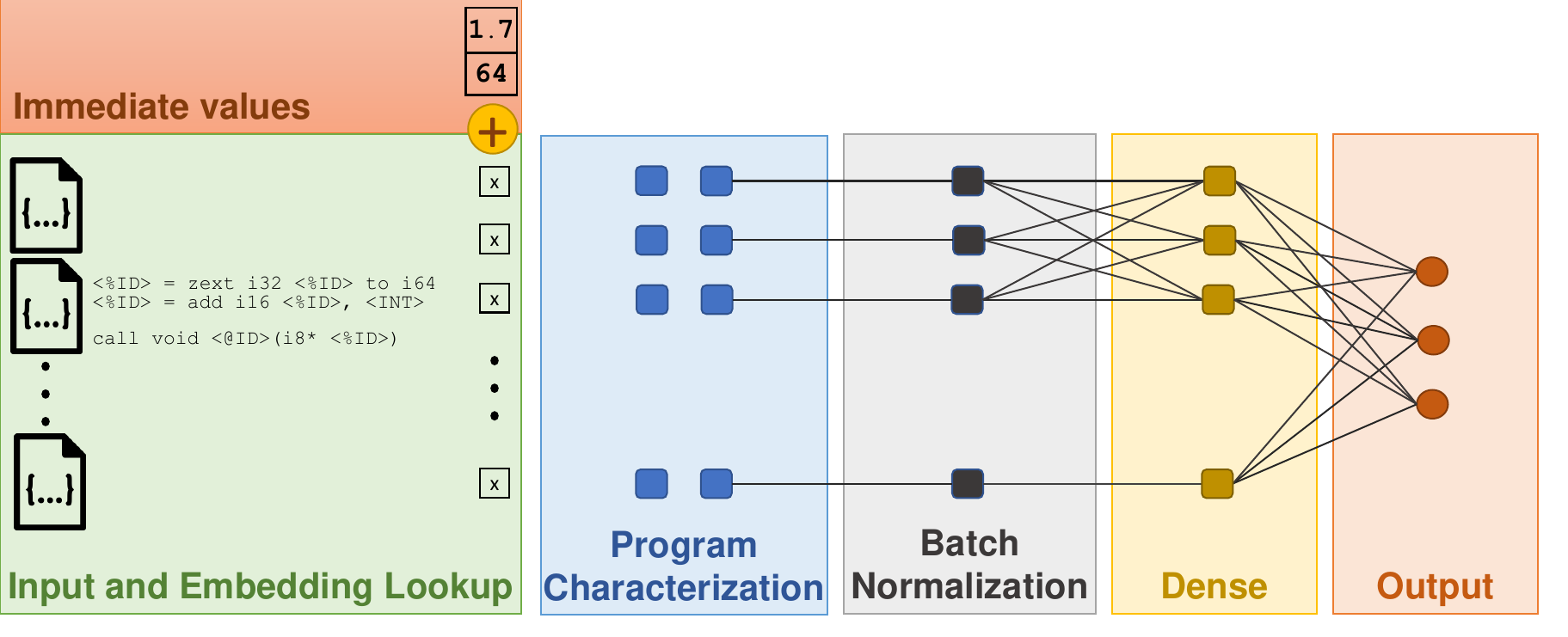}
		\caption{\texttt{concat\_naïve}}
		\label{fig:nccimm:concat_naive}
	\end{subfigure}
	\begin{subfigure}[t]{\textwidth}
		\centering
		\includegraphics[page=3,height=1.3in,trim={0cm 6.5cm 0cm 0cm},clip]{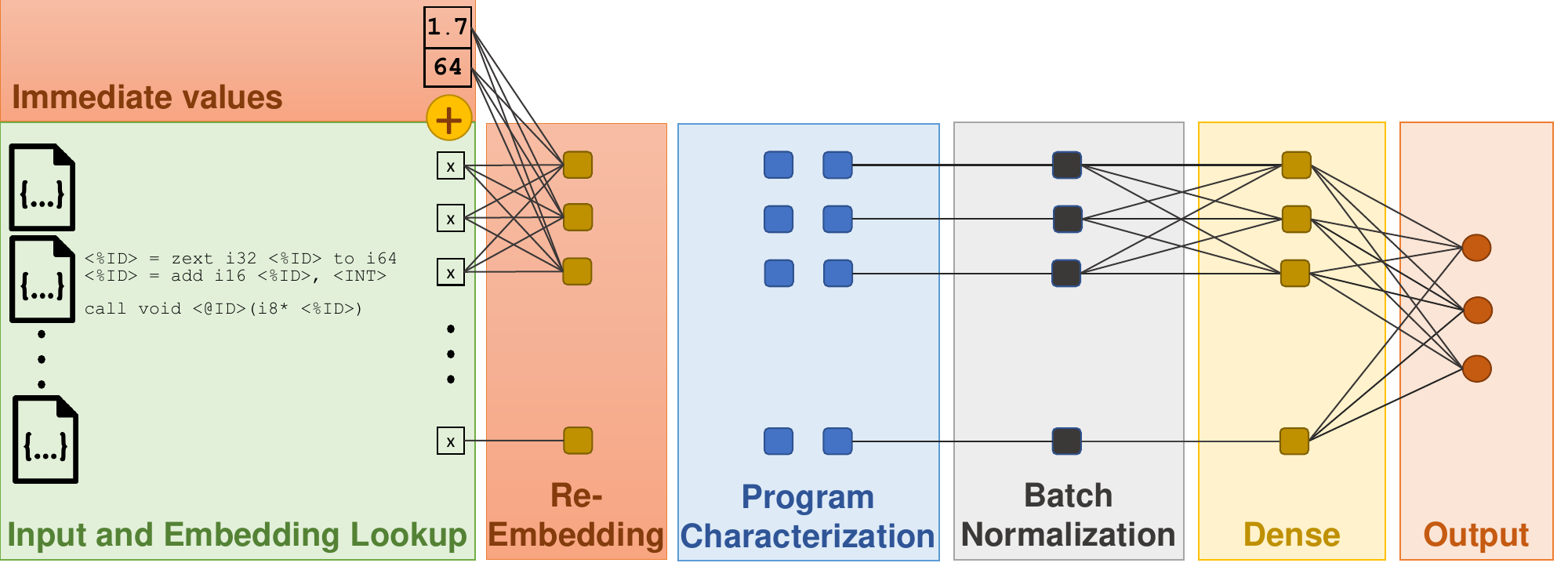}	
		\caption{\texttt{concat\_embed}}
		\label{fig:nccimm:concat_emb}
	\end{subfigure}
	\begin{subfigure}[t]{\textwidth}
		\centering
		\includegraphics[page=4,height=1.3in,trim={0cm 6.5cm 3cm 0cm},clip]{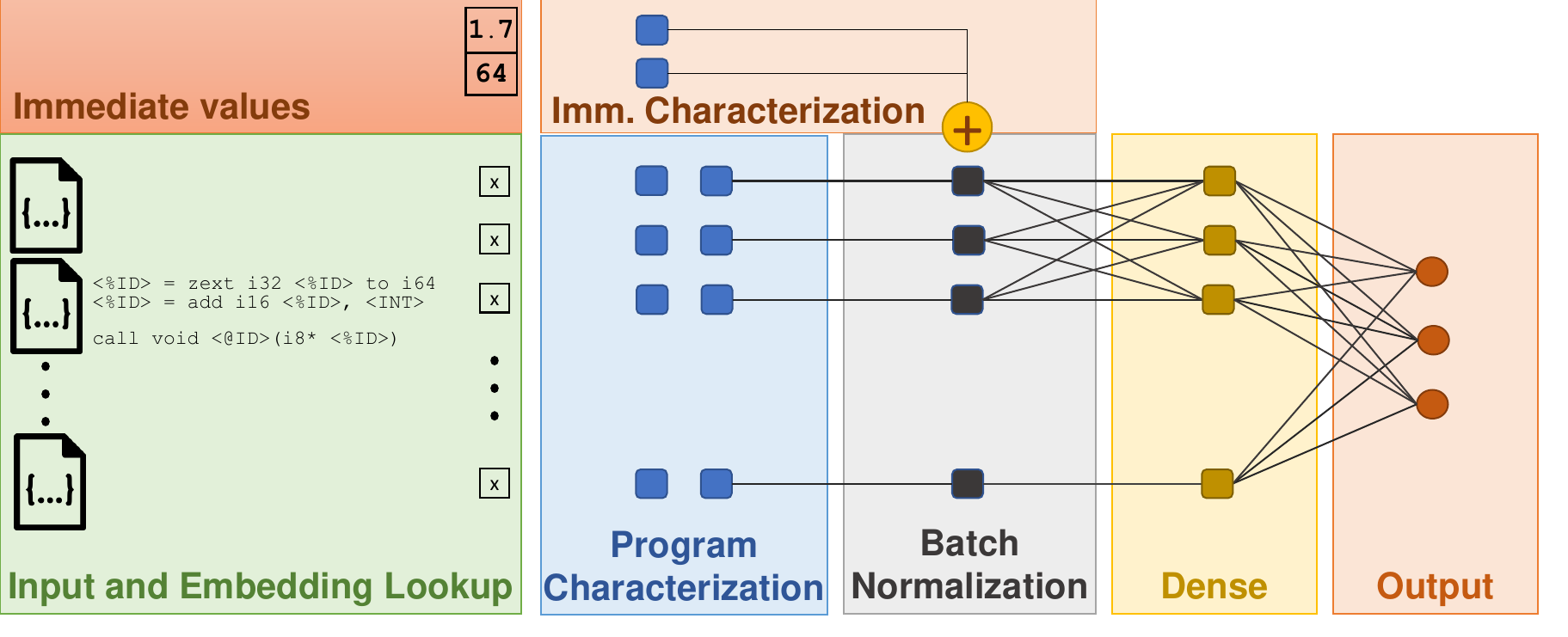}
		\caption{\texttt{extract\_concat}}
		\label{fig:nccimm:extract_concat}
	\end{subfigure}

	\caption{Three architectures for training \texttt{inst2vec} sequences of statements along with their immediate values in NCC. The components related to immediate values are marked in dark orange. The stage at which the immediate values are concatenated with the statements is denoted with a yellow ``$+$''~sign.}
	\label{fig:nccimm}
\end{figure}

In the transformations applied to raw LLVM IR code before \texttt{inst2vec} training, the statements are stripped of their immediate values and are replaced by tokens indicating the value type: \texttt{<INT>}, \texttt{<FLOAT>}, \texttt{<STRING>} (see Section \ref{sec:inst2vec} for further detail). The purpose of this abstraction from the immediate values is twofold. First, keeping all immediate values would result in an extremely large and sparse vocabulary size; second, this transformation allows to map statements with nearly identical semantics (e.g. \texttt{<\%ID> = fadd fast float <\%ID>, 1.3} and \texttt{<\%ID> = fadd fast float <\%ID>, 5.2}) to the same embedding vector. While this choice is sound for statement training, it might nevertheless fall short in the training of downstream tasks, where immediate values may hold values critical to the program's semantics or performance, such as array sizes or iteration bounds. In order to train \texttt{inst2vec} program representations along with their immediate values, we store the immediate values of each statement separately before filtering them out during preprocessing. The immediate values are then reintegrated into the NCC workflow using one of the three methods illustrated in Fig.~\ref{fig:nccimm} and described below.

\textbf{``naïve concatenation'' (\texttt{concat\_naïve})}~~~Instead of feeding the 
model with the embedding vector of a statement alone (see layer~$2$, above), embedding vectors are first concatenated with their corresponding immediate values. The first set of LSTM cells accept an input of size \textit{embedding dimension + length of list of immediates}. The remainder of the NCC layers are unchanged.

\textbf{``concatenate then embed'' (\texttt{concat\_embed})}~~~This method introduces an additional embedding step: statement embedding vectors are first concatenated with their corresponding immediate values. They then pass through a fully-connected layer, which reduces the layer dimension from \textit{input dimension = embedding dimension + length of list of immediates} back to \textit{embedding dimension}. Next, program characterization and the rest of the network remain unchanged.

\textbf{``extract then concatenate'' (\texttt{extract\_concat})}~~~In this method, immediate values are never coupled back directly to the statement from which they were extracted. Rather, the sequence of immediate values of the entire program undergoes a separate processing pipeline, before being concatenated with the output of the program characterization as auxiliary inputs (see Fig. \ref{fig:ncc:arch}). The separate processing of immediate values sequence consists of a single LSTM layer, designed to extract the critical information from the immediate values that characterize the program.
	
\newpage
\section{Training NCC with Immediate Values: Exhaustive Results}
\label{app:res:immval}

Tables~\ref{res:task:devmap:immval} and~\ref{res:task:cf:compare_speedups:immval} present the results for the heterogeneous device mapping and optimal thread coarsening factor tasks, obtained with the different modes of immediate value handling described in Appendix~\ref{app:method:immval}. The column 'ignore' presents the results for the simplest version of NCC, where immediate values are ignored.

\begin{table}[h]
	\caption{Heterogeneous device mapping results obtained with \texttt{inst2vec} and NCC, using different modes of immediate value handling}
	\label{res:task:devmap:immval}
	\centering
	\scriptsize
	\begin{tabular}{lcccccccc}
		\toprule
		Architecture        & \multicolumn{4}{c}{Prediction Accuracy [\%]} & \multicolumn{4}{c}{Speedup}\\
		\cmidrule(r){2-5}  \cmidrule(r){6-9}
		& ignore & concat  & extract & concat & ignore & concat  & extract & concat \\
		& & naïve & concat & emb & & naïve & concat & emb \\
		\midrule
		AMD Tahiti 7970     & 82.79 & \textbf{88.09} & 76.18 & 72.06
		& 3.42 & \textbf{3.47} & 3.36 & 2.76 \\
		NVIDIA GTX 970      & 82.06 & \textbf{86.62} & 79.71 & 72.50 
		& 1.42  & \textbf{1.44} & 1.40 & 1.32\\
		\bottomrule
	\end{tabular}
\end{table}

\begin{table}[h]
	\caption{Speedups achieved by coarsening threads with \texttt{inst2vec} and NCC, using different modes of immediate value handling}
	\label{res:task:cf:compare_speedups:immval}
	\centering
	\scriptsize
	\begin{tabular}{lcccc}
		\toprule
		Computing Platform        & \multicolumn{4}{c}{Speedup}\\
		\cmidrule(r){2-5}
		& ignore & concat\_naïve  & extract\_concat & concat\_embed \\
		\midrule
		AMD Radeon HD 5900  & \textbf{1.37}	& 1.21 & 1.28 			& 1.30 \\
		AMD Tahiti 7970     & 1.10 			& 1.06 & \textbf{1.18}  & 0.92 \\
		NVIDIA GTX 480      & 1.07 			& 0.99 & \textbf{1.11}  & 0.97 \\
		NVIDIA Tesla K20c   & \textbf{1.06} & 1.04 & 1.00 		    & 0.99 \\
		\bottomrule
	\end{tabular}
\end{table}

\end{document}